\documentclass{article}

 \PassOptionsToPackage{numbers,compress}{natbib}
 \usepackage[dblblindworkshop,final]{neurips_2025}

 \usepackage{graphicx}

\usepackage[utf8]{inputenc} 
\usepackage[T1]{fontenc}    
\usepackage{hyperref}       
\usepackage{url}            
\usepackage{booktabs}       
\usepackage{amsfonts}       
\usepackage{nicefrac}       
\usepackage{microtype}      
\usepackage{xcolor}         

\title{Demo: Guide-RAG: Evidence-Driven Corpus Curation for Retrieval-Augmented Generation in Long COVID}
\workshoptitle{The Second Workshop on GenAI for Health: Potential, Trust, and Policy Compliance}
\author{%
  Philip DiGiacomo \\
  Department of Computer Science\\
  University of Texas at Austin\\
  \And
  Haoyang Wang \\
  School of Information\\
  University of Texas at Austin\\
  \And
  Jinrui Fang \\
  School of Information\\
  University of Texas at Austin\\
  \And
  Yan Leng \\
  McCombs School of Business\\
  University of Texas at Austin\\
  \And
  W Michael Brode \\
  Dell Medical School\\
  University of Texas at Austin\\
  \And
  Ying Ding \\
  School of Information\\
  University of Texas at Austin\\
}

\begin{document}

\maketitle{}

\begin{abstract}
As AI chatbots gain adoption in clinical medicine, developing effective frameworks for complex, emerging diseases presents significant challenges. We developed and evaluated six Retrieval-Augmented Generation (RAG) corpus configurations for Long COVID (LC) clinical question answering, ranging from expert-curated sources to large-scale literature databases. Our evaluation employed an LLM-as-a-judge framework across faithfulness, relevance, and comprehensiveness metrics using LongCOVID-CQ, a novel dataset of expert-generated clinical questions. Our RAG corpus configuration combining clinical guidelines with high-quality systematic reviews consistently outperformed both narrow single-guideline approaches and large-scale literature databases. Our findings suggest that for emerging diseases, retrieval grounded in curated secondary reviews provides an optimal balance between narrow consensus documents and unfiltered primary literature, supporting clinical decision-making while avoiding information overload and oversimplified guidance. We propose Guide-RAG, a chatbot system and accompanying evaluation framework that integrates both curated expert knowledge and comprehensive literature databases to effectively answer LC clinical questions.
\end{abstract}

\section{Introduction}
Artificial intelligence (AI) tools and large language models (LLMs) are rapidly being adopted in clinical medicine, with both patients and clinicians increasingly turning to chatbot platforms for on-demand medical information \cite{wang2024applications, thirunavukarasu2023large, huo2025large, goodman2023accuracy, bedi2025testing, yun2025online, laymouna2024roles}. For clinicians, resources such as OpenEvidence are emerging to augment traditional references like Harrison’s textbook or UpToDate online reference, offering algorithmically synthesized summaries of clinical literature \cite{hurt2025use,prorok2012quality}. This rapid uptake underscores both the promise and the risk of chatbot-based decision support: while these systems can provide interactive, timely guidance, they remain vulnerable to hallucination, citation errors, and representational biases \cite{wang2024applications,huo2025large,goodman2023accuracy,bedi2025testing,bernstein2023comparison}. Designing best-practice frameworks for medical chatbots has therefore become an urgent challenge, especially for diseases that are complex, heterogeneous, or poorly understood \cite{huo2025large,bedi2025testing,huo2025reporting}.

Long COVID (LC) represents the challenge of treating complex and emerging illnesses. Affecting an estimated 7\% of U.S. adults (approximately 18 million people), LC is a heterogeneous, multisystem condition characterized by more than 200 reported symptoms, and lacks a standardized diagnostic biomarker or evidence-based treatment \cite{national2024long,chou2024long,ely2024long}. Clinical guidance remains largely consensus-based and is often extrapolated from related syndromes such as myalgic encephalomyelitis/chronic fatigue syndrome (ME/CFS) and postural orthostatic tachycardia syndrome (POTS) \cite{brode2024practical,cheng2025multidisciplinary,al2024long}. In such high-uncertainty settings without clear ground truth, designing chatbots for clinical decision support is particularly challenging, as relying on PubMed alone or a single guideline risks overwhelming clinicians or missing evolving evidence. Retrieval-Augmented Generation (RAG) systems offer a promising approach to address these limitations by grounding language model responses in external evidence. The selection of corpus documents is critical to the system’s capability to answer clinical questions. In the case of emerging diseases, current approaches span two extremes: manually curated, high-quality databases versus large-scale literature collections that provide comprehensive coverage but introduce quality and relevance challenges. We propose a balanced approach using a novel combination of a LC consensus guideline with high–quality systematic reviews.

We evaluated RAG performance across three dimensions: faithfulness (ensuring responses are grounded in source documents), relevance (maintaining focus on the clinical question), and comprehensiveness (providing thorough coverage of complex, multisystem presentations). Our evaluation employed an LLM-as-a-judge framework that went beyond traditional accuracy metrics to assess criteria directly relevant to clinical practice. Additionally, we developed LongCOVID-CQ, a specialized dataset of expert-generated clinical questions targeting diagnosis, management strategies, and mechanisms that practicing clinicians routinely encounter when caring for LC patients. We summarize Guide-RAG's three key contributions:
\begin{enumerate}
    \item \textbf{Expert corpus curation}: Targeted curation of a clinical guideline supplemented by three high-quality systematic reviews outperformed both large-scale literature databases and narrow single-guideline approaches for LC question answering using RAG.
    \item \textbf{Evaluation metrics}: We adopted faithfulness, relevance, and comprehensiveness metrics specifically for LC clinical applications using an LLM-as-a-judge framework to capture criteria directly relevant to clinical decision-making and trust-building.
    \item \textbf{LongCOVID-CQ}: We developed a specialized evaluation dataset of expert-generated long COVID clinical questions (LongCOVID-CQ) reflecting the practical information needs that providers routinely encounter in patient care.
\end{enumerate}
\section{Related Work}
\textbf{RAG applications to emerging and unknown diseases.} Emerging diseases present a dual challenge for information systems: the scarcity of high-quality data \cite{metcalf2017opportunities} and the overload of rapidly published yet conflicting information \cite{zarocostas2020fight}. Compact, domain-specific knowledge bases can improve RAG performance via factual recall and diagnostic reasoning \cite{song2025graph,zelin2024rare}. However, quickly evolving research may not be captured by the curated knowledge base. Using large but unfiltered sources such as PubMed presents distinct challenges \cite{li2025use}: irrelevant or low-quality articles can degrade model accuracy, and the “lost-in-the-middle” effect can prevent models from recalling information buried within extensive text. Clinical guidelines are useful tools in navigating the specific versus general knowledge tradeoff, by distilling clinical consensus out of broad research topics. Guidelines for well-studied medical subjects have been successively applied to RAG pipelines in contrast-media consultations and perioperative assessments \cite{wada2025retrieval,ke2025retrieval}. We found that combining guidelines with systematic reviews widened coverage without re-introducing noise. To our knowledge, no published study has evaluated a RAG framework that grounds answers jointly in clinical guidelines and systematic reviews for LC or other emerging and unknown diseases.

\textbf{Evaluation of RAG for clinical relevance.} Existing medical question-answering (QA) benchmarks predominantly employ multiple-choice formats \cite{jin2021disease, pal2022medmcqa, jin2019pubmedqa, hendrycks2020measuring}. Consequently, medical-specific RAG chatbots are evaluated primarily on accuracy metrics when selecting from predetermined options \cite{nori2023can, xiong2024benchmarking, RevLLMClinicalMed}. This narrow focus on accuracy fails to capture critical aspects of clinical utility, including decision support capabilities and adherence to clinical guidelines \cite{MedHELM, EvalMitigateLLMClinical}. Recent work has begun exploring multi-dimensional evaluation of RAG systems in clinical settings using LLM-as-a-judge \cite{zheng2023judging}. \citet{wada2025retrieval} evaluated response time, applicability, structure, safety, and professional communication, while \citet{MedGraphRAG} assessed pertinence, correctness, understandability, precision, and recall. However, the abundance of available metrics necessitates principled selection based on downstream application requirements. For LC clinical application, we identify three evaluation dimensions. First, faithfulness ensures clinician trust by requiring responses to be substantiated by source documents \cite{es2024ragas}. Second, given the 200+ reported symptoms of LC \cite{national2024long,chou2024long,ely2024long}, chatbots must maintain relevance to presented symptoms \cite{es2024ragas}. Finally, comprehensiveness ensures holistic assessment across the vast symptom space \cite{edge2024local}. We utilized an LLM-as-a-judge evaluation framework that assessed RAG systems across faithfulness, relevance, and comprehensiveness for LC queries. This approach went beyond accuracy-based metrics while targeting criteria essential for LC clinical practice.
\begin{figure}[t]
    \centering
    \includegraphics[width=\textwidth]{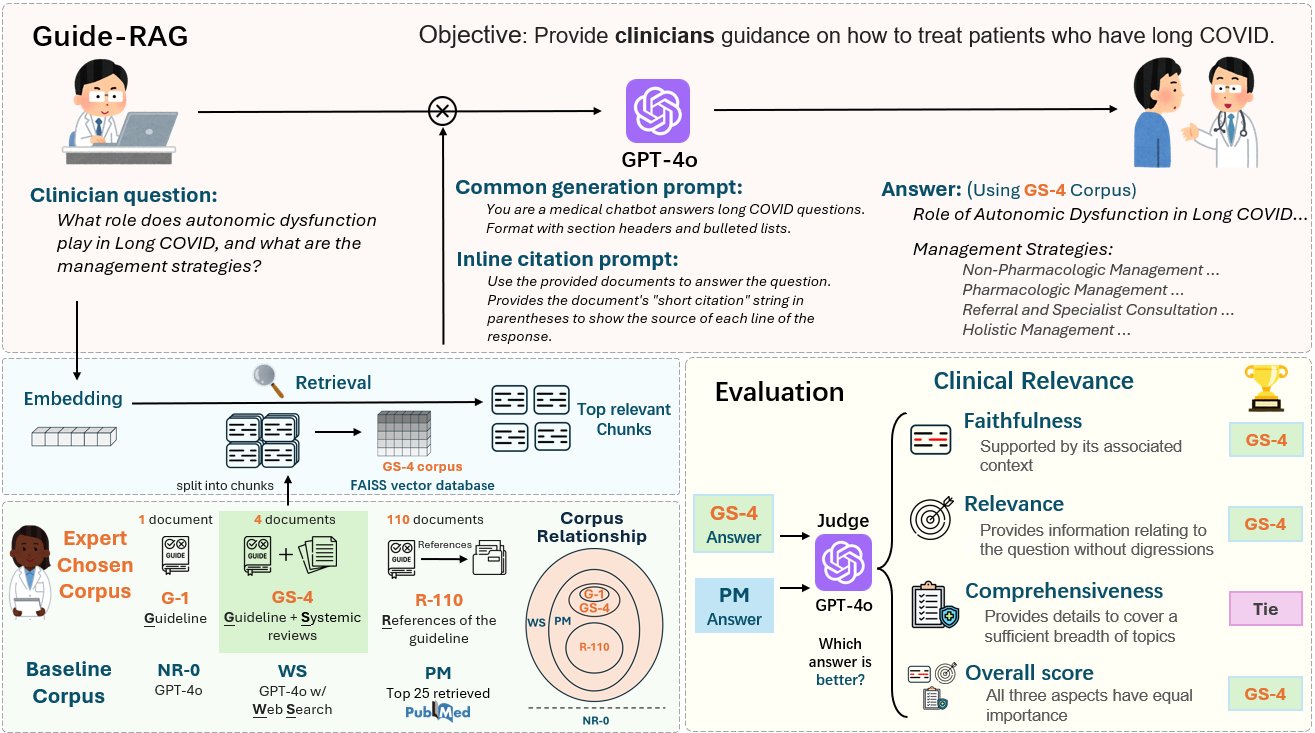}
    \caption{Workflow of Guide-RAG answering LC questions. The full prompt used for the answer generation and evaluation is presented in Appendix A.}
    \label{fig:workflow}
\end{figure}
\section{Methods}
\label{gen_inst}
\subsection{Corpus construction}
We systematically evaluated six distinct corpus configurations within our Guide-RAG framework to assess the impact of document curation strategies on clinical question-answering performance (Figure~\ref{fig:workflow}). We performed our analysis across faithfulness, relevance, and comprehensiveness metrics, while including an overall metric that combines all three metrics equally. Our experimental design contrasts expert-curated knowledge bases with large-scale literature corpuses using both sparse and dense retrieval methods (detailed in Appendix C). This investigation addressed the clinical challenge where existing frameworks relying on vast, unfiltered literature repositories like PubMed may produce technically accurate but clinically misaligned outputs, particularly for emerging diseases with limited diagnostic clarity and therapeutic consensus.

\textbf{No retrieval (NR-0).} Standard GPT-4o without retrieval augmentation served as our control condition.

\textbf{Expert-curated corpuses.} Three configurations leveraged domain expertise from a physician-scientist specializing in LC care. \textbf{Guideline only (G-1).} This minimal configuration consisted solely of the American Academy of Physical Medicine and Rehabilitation (AAPM\&R) "Multidisciplinary collaborative guidance on the assessment and treatment of patients with Long COVID: A compendium statement" \cite{cheng2025multidisciplinary}. This document represents the current consensus-driven guidance on LC in the United States, developed through a multi-year collaborative process by active researchers and clinicians. \textbf{Guideline + systematic reviews (GS-4).} This configuration augmented G-1 with three high-quality systematic reviews selected for comprehensiveness and recency, providing the current synthesis of the evolving LC literature from leading medical journals. \textbf{References of the guideline (R-110).} This corpus comprised the 110 references cited within the AAPM\&R guideline. We excluded one overlapping citation with GS-4 to ensure independent comparison \cite{national2024long} and two references were omitted due to the infeasibility of preprocessing as detailed in Appendix H. This configuration tested whether grounding the model in the evidence base underlying consensus guidance improved response quality relative to summary documents alone.

\textbf{Large-scale literature corpuses.} We adopted vast bodies of available research as baselines for comparison. \textbf{PubMed corpus (PM).} This configuration accessed the comprehensive biomedical literature database maintained by the National Library of Medicine (NLM), encompassing MEDLINE (>39M citations), PubMed Central (full-text articles), and Bookshelf (biomedical books). The scale necessitated a hybrid sparse-dense retrieval approach (Appendix C). \textbf{Web search (WS).} This configuration employed OpenAI's GPT-4o web search capabilities with explicit constraints to retrieve only peer-reviewed medical publications with source attribution.

\subsection{Dataset: LongCOVID-CQ}
A physician-scientist specializing in LC care developed 20 questions reflecting the key issues a practicing clinician is likely to encounter. Rather than a general summary, these queries target clinically relevant topics such as diagnosis, management strategies, and mechanisms to capture the kinds of questions providers routinely face in caring for patients with LC. These questions are listed in Appendix B.
\subsection{Evaluation framework}
\textbf{Evaluation metrics.} We employed three criteria, plus an overall metric, targeting responses that are useful for clinical application. \textbf{Faithfulness} measures whether generated content is directly supported by retrieved document chunks, enabling citation traceability essential for clinical trust and verification. \textbf{Relevance} quantifies the degree to which responses directly address the input question without extraneous content that burdens clinical workflow. \textbf{Comprehensiveness} evaluates coverage breadth, which is important given the >200 documented LC symptoms requiring multi-faceted responses. \textbf{Overall} performance aggregates faithfulness, relevance, and comprehensiveness with equal weighting to provide holistic quality assessment.

\textbf{LLM-as-a-Judge} We employed GPT-4o as an evaluation model for pairwise comparisons across all corpus configurations. For each input question, responses were generated from all six configurations and evaluated head-to-head across the three criteria mentioned, plus an overall metric. The evaluation prompt included explicit instructions for handling response length bias and positional effects, with randomized presentation order to mitigate systematic biases. All evaluation prompts included one-shot examples and explicit tie-handling instructions. Complete prompting details are provided in Appendix A. Win rates for each comparison were calculated by assigning a score of 100 to the winning response, 0 to the losing response, and 50 to both responses in case of a tie. These scores were then averaged over the questions in LongCOVID-CQ.
\begin{figure}[t]
    \centering
    \includegraphics[width=1\linewidth]{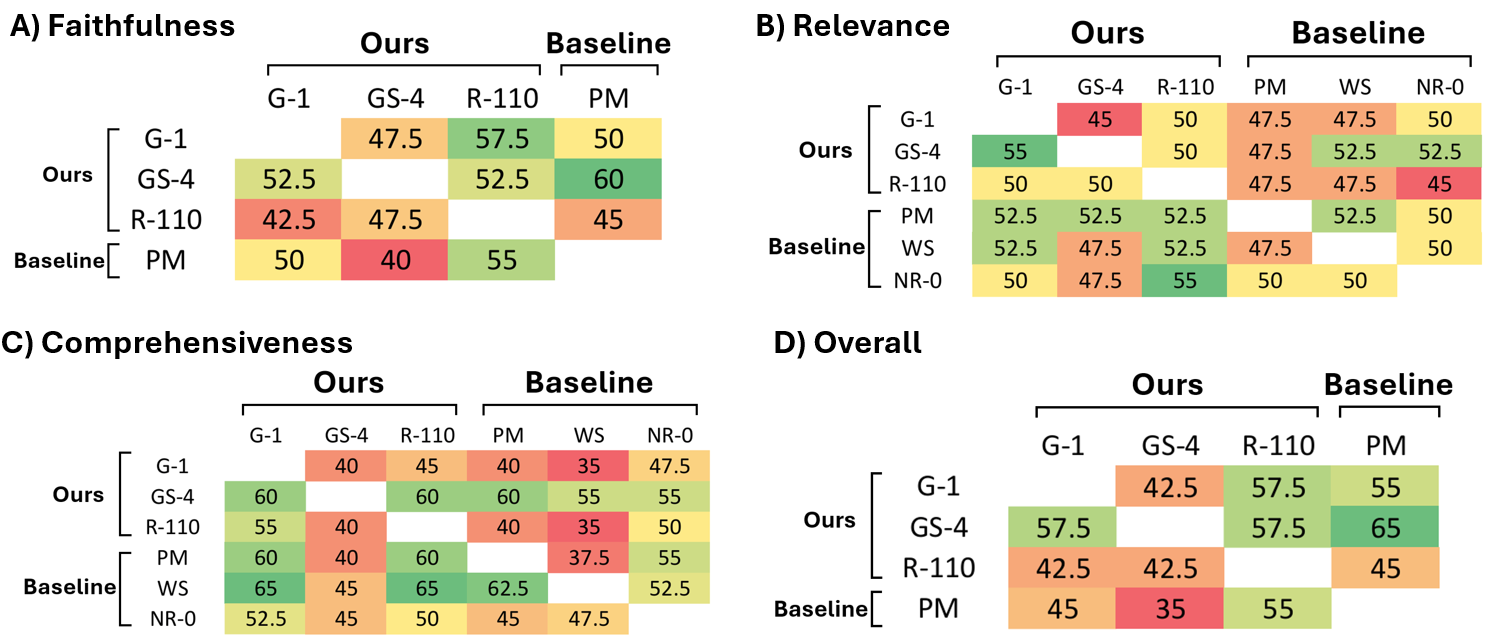}
    \caption{Win rates for (A) faithfulness, (B) relevance, (C) comprehensiveness, and (D) overall (each cell is the percentage of comparisons 
    won by the row label over the column label)}
    \label{fig:heatmaps}
\end{figure}
\section{Results}
\label{Results}
\textbf{GS-4 achieved superior overall performance.} Our overall evaluation encompassing all three metrics demonstrated that GS-4 consistently outperformed competing approaches, achieving win rates of 57.5-65\% in pairwise comparisons. This superior performance extended to two of the three individual metrics, with GS-4 ranking highest in both faithfulness and comprehensiveness evaluations. Notably, despite utilizing only 4 sources, GS-4 achieved greater comprehensiveness than both R-110 and PM (60\% win rate for both comparisons), which leveraged significantly larger corpuses. Win rate data is shown for all comparisons in Figure~\ref{fig:heatmaps}. Our heatmaps did not include NR-0 or WS for the faithfulness or overall comparisons because retrieved chunks of text were required to make those comparisons.

\textbf{PM exhibited strongest relevance performance.} While GS-4 dominated overall metrics, PM demonstrated slight superiority in relevance evaluation, achieving win rates of 50-52.5\% against other conditions. We hypothesize that PM's access to expanded information enables this enhanced relevance performance.

\textbf{Guideline outperformed constituent references.} The comparison between G-1 and R-110 provided insight into the relative efficacy of synthesized guidelines versus their underlying references. G-1 achieved a 57.5\% win rate over R-110 in faithfulness and overall evaluation. While R-110 maintained an advantage in comprehensiveness (55\% win rate) and had comparable relevance performance (50\% win rate), G-1's overall superiority suggests that synthesized guidelines can provide enhanced faithfulness through better alignment with retrieved content, while being comprehensive and relevant enough to not degrade overall performance.

\textbf{Expert Evaluation.} Beyond the quantitative win-rate data, expert clinical review helps contextualize these results for real-world LC care. The heatmaps showed that corpora built from expert-curated sources consistently outperformed broader or unstructured collections, especially in domains like symptom management and mechanisms where overly specific or overconfident responses can be misleading. Figure~\ref{fig:Examples1} illustrates this with side-by-side outputs from different corpus conditions. Broader corpora (R-110 and PM) produced answers that appeared superficially plausible in their specificity but leaned heavily on single studies, and overemphasized speculative mechanisms. For example, R-110 attributed autonomic dysfunction solely to vagus nerve imbalance, and even recommended poorly researched interventions such as stellate ganglion block. The PM response went further, endorsing exercise training based on a review article that extrapolated treatment approaches from other diseases—advice that is explicitly cautioned against by LC experts, where standard exercise regimens may worsen symptoms. By contrast, the curated GS-4 corpus framed autonomic dysfunction as one part of a systemic illness and offered practical management guidance that acknowledged evidence gaps and suggested specialist involvement when appropriate. It also correctly noted that any physical activity must be carefully tailored to the patient. From a clinical perspective, this response is far more useful: it equips clinicians to counsel patients responsibly while avoiding misleading certainty. This example underscores how uncurated data can generate outputs that appear authoritative yet are incomplete or misleading, and why expert evaluation remains essential alongside quantitative benchmarks when developing AI systems for emerging illnesses like LC. Full responses can be compared in Appendix D-F.
\section{Discussion and Conclusion}
These findings suggest a design principle for chatbots in emerging disease contexts: retrieval grounded in curated secondary reviews may provide a balance between narrow consensus documents and the unfiltered breadth of primary literature. The early COVID-19 pandemic illustrates the stakes, with more than 23,000 COVID-19 articles indexed in the first six months alone, many offering little new or meaningful information \cite{teixeira2021publishing}. In such settings, indiscriminate retrieval risks producing technically accurate but clinically misaligned outputs. Our evaluation provides preliminary evidence that systematic reviews and expert guidance can anchor chatbot responses in a way that supports faithfulness, relevance, and comprehensiveness: qualities that go beyond accuracy and are essential for decision support in high-uncertainty medical domains.
The LongCOVID-CQ dataset addresses a critical gap in medical AI evaluation by providing clinically grounded questions that reflect real-world information needs. Unlike existing medical QA benchmarks that predominantly employ multiple-choice formats testing factual recall, our expert-generated questions target the diagnostic uncertainty, management complexity, and mechanistic understanding that practicing clinicians actually encounter when caring for LC patients. 
\section{Limitations}
This study was limited by its focus on a single clinical domain (LC) and by the use of a single LLM model (GPT-4o) as the evaluation judge without validation against human or model-diverse raters. The small, expert-generated question set limited statistical robustness, and differences in retrieval strategies across corpora, such as dense retrieval for smaller datasets versus hybrid sparse-dense retrieval for PubMed, may have introduced bias in recall quality. The study also did not include systematic exploration of retrieval hyperparameters, including chunk size, embedding model selection, and reranking thresholds. These factors together constrain the generalizability and reproducibility of the findings.

Future work should include human expert ratings, multiple model evaluations, retrieval ablations, and will extend testing to additional high-complexity and high-uncertainty medical domains. By beginning with LC, a condition characterized by high-complexity and evolving evidence, this framework may provide generalizable principles for developing trustworthy AI systems across other complex diseases.
\small           
\bibliographystyle{unsrtnat} 
\bibliography{references}


\appendix

\section*{Appendix A: Prompts}
\subsection*{Common generation prompt}
The following prompt is used for generation of answers in all corpus selections. 

\begin{quote}
You are a medical chatbot that is answering a long COVID question asked by a clinician. Format your response using markdown with section headers and bulleted lists.
\end{quote}

\subsection*{Inline citation prompt}
The following prompt is used to generate a response to the user question based on the document chunks retrieved. This only applies to the G-1, GS-4, R-110, and PM modes.

\begin{quote}
Use the provided documents to answer the question. Create inline citations by providing the document's "short\_citation" string surrounded by parentheses to show exactly where each line of the response originates from. Only incorporate evidence from the provided documents that is relevant to answering the question.
\end{quote}

\subsection*{Web search prompt}
The following prompt is used for GPT-4o when web search is enabled.

\begin{quote}
You must reply using information from peer reviewed medical publications only. Please cite your sources.
\end{quote}

\subsection*{PubMed Query Prompt}
The following prompt takes the user question and creates a PubMed query using a single-shot example.

\begin{quote}
You are a PubMed search expert. Create a PubMed advanced search query for the following question:

Question: \{question\}

Requirements:

1. MUST include 'AND "long COVID"' in the query

2. Use PubMed advanced search syntax with parentheses, AND, OR operators

3. Return only the search query, nothing else

Examples:

Question: "What are the specific neurological symptoms of long COVID in elderly patients?"

Query: (neurological OR neurology) AND (elderly OR seniors) AND "long COVID"
\end{quote}

\subsection*{PubMed “Rewrite Query” Prompt}
When insufficient results are returned, this prompt is used to create a new PubMed search query that seeks to elicit more results than the previous query.

\begin{quote}
You are a PubMed search expert. The previous search query failed to find enough results. Create a more relaxed PubMed search query based on the original question.

Original Question: \{question\}

Failed Query: \{failed\_query\}

Requirements for the relaxed query:

1. MUST include 'AND "long COVID"' in the query

2. Use more general terms and broader synonyms than the failed query

3. If the terms cannot be made more general, use fewer terms
\end{quote}

\subsection*{Pairwise Evaluation Prompt}

The following prompt compares two responses head-to-head based on a given criteria. The “Context for Answer 1” and “Context for Answer 2” were included when evaluating faithfulness and overall. Context was not included when evaluating relevance and comprehensiveness. This prompt was adapted from \citet{edge2024local} and \citet{zheng2023judging}. 

\begin{quote}
—Goal—
Given a question and two answers (Answer 1 and Answer 2), assess which answer is better according to the following measure:
\{criteria\_text\}

Your assessment should include two parts (in the given order):

Reasoning: a short explanation of why you chose the winner with respect to the measure described above.

Winner: this should be 1 (if Answer 1 is better), 2 (if Answer 2 is better), or 0 if they are fundamentally similar and the differences are immaterial.

Ensure that the order in which the responses were presented does not influence your decision. Do not allow the length of the responses to influence your decision.

Question: \{question\}

Context for Answer 1: 
\{retrieved chunks for answer 1\}
           
Context for Answer 2:
\{retrieved chunks for answer 2\}

Answer 1: \{answer 1\}

Answer 2: \{answer  2\}

Assessment:
\end{quote}

\subsection*{Output format prompt}
The following was used with the pairwise evaluation prompt to ensure the output would be json that could be parsed.

\begin{quote}
IMPORTANT: Respond with ONLY a valid JSON object in exactly this format:

\{``comparisons'': [\{``reasoning'': ``explanation for this specific comparison'',``winner'': Z, \}, ...]\}

Where:

- reasoning is a short explanation specific to this pairwise comparison

- Z is the winner (1, 2, or 0 for tie)

Do not include any text before or after the JSON. Do not use markdown formatting or code blocks.
\end{quote}

\subsection*{Faithfulness criteria}
The following was adapted from \citet{edge2024local} and used with the pairwise evaluation prompt as the criteria for faithfulness.

\begin{quote}
Is the answer supported by its associated context? If the answer contains claims that cannot be substantiated by the retrieved context, it is unfaithful, even if it is factually correct. For example, the question is “What are the benefits and drawbacks of nuclear energy”, and the retrieved context involves information from a study that shows radiation released during the Three Mile Island accident is related to increased cancer incidence in the area. A faithful answer would cite the Three Mile Island accident to warn of its potential danger. An answer citing the Chernobyl accident, without being present in the retrieved context, would be unfaithful. An answer that incorrectly uses the Three Mile Island accident to say that nuclear energy does not pose potential risks would be unfaithful.
\end{quote}

\subsection*{Relevance criteria}
The following  was adapted from \citet{edge2024local} and used with the pairwise evaluation prompt as the criteria for relevance.

\begin{quote}
Does the answer provide information that relates to the question? If the answer contains digressions from topic established by the question, it would demonstrate irrelevance. For example, if the question is “What are the benefits and drawbacks of nuclear energy”, a relevant answer may discuss the large number of kilowatts it produces annually, as well as the risks posed by a nuclear meltdown. An answer that discusses the danger of a nuclear bomb, without connecting this to nuclear power generation, would be irrelevant.
\end{quote}

\subsection*{Comprehensiveness criteria}
The following was adapted from \citet{edge2024local} and used with the pairwise evaluation prompt as the criteria for comprehensiveness.

\begin{quote}
How much detail does the answer provide to cover all the aspects of the question? A comprehensive answer should be thorough and complete. For example, if the question is “What are the benefits and drawbacks of nuclear energy”, a comprehensive answer would provide both the positive and negative aspects of nuclear energy, such as its efficiency, environmental impact, safety, cost, etc. A comprehensive answer should not leave out any important points. For example, an incomplete answer would only provide the benefits of nuclear energy without describing the drawbacks.
\end{quote}

\subsection*{Overall criteria}
The following was used with the pairwise evaluation prompt as the criteria for the overall metric.

\begin{quote}
Each of the following criteria should be weighted equally.
Faithfulness: the answer should be supported by its associated context. If the answer contains claims that cannot be substantiated by the retrieved context, it is unfaithful, even if it is factually correct.
Comprehensiveness: the answer should provide details to cover all the aspects of the question.
Relevance: the answer should provide information that relates to the question without digressions.

\end{quote}

\section*{Appendix B: LongCOVID-CQ dataset}

\subsection*{General Knowledge and Evidence Synthesis}
\begin{enumerate}
    \item What are the most common symptoms of Long COVID, and how do they vary between patients?
    \item Can you summarize the current evidence on the pathophysiology of Long COVID?
    \item What role does autonomic dysfunction play in Long COVID, and what are the management strategies?
    \item What evidence-based therapies exist for treating fatigue in Long COVID?
    \item What does the current evidence say about functional impairment and disability in patients with Long COVID, and how can patients get disability protections?
\end{enumerate}

\subsection*{Clinical Scenarios and Decision Support}
\begin{enumerate}
    \item A 45-year-old patient presents with persistent shortness of breath 8 months after mild COVID-19. What tests would you recommend and why?
    \item How would you differentiate Long COVID-related cognitive dysfunction from other causes of cognitive impairment?
    \item What treatment options would you recommend for a patient with POTS (postural orthostatic tachycardia syndrome) related to Long COVID?
    \item A patient with Long COVID presents with significant mental health symptoms. What screening tools and management steps should be considered?
    \item How would you approach rehabilitation for a patient with Long COVID-related post-exertional malaise (PEM)?
    \item A 72 year-old black woman with pre-existing hypothyroidism and fibromyalgia presents with severe fatigue, muscle aches, chest pain, and abdominal pain following COVID-19 infection in July 2022. What tests should be ordered, and what treatment should she receive?
\end{enumerate}

\subsection*{Controversies and Evolving Evidence}
\begin{enumerate}
    \setcounter{enumi}{11}
    \item What is the current evidence for antiviral therapy or immunomodulators in managing Long COVID?
    \item How reliable is the use of biomarkers, such as CRP or cytokines, in diagnosing or monitoring Long COVID?
    \item What does the evidence say about the impact of COVID-19 vaccination on Long COVID symptoms?
    \item Is there any evidence to support dietary interventions or supplements for Long COVID management?
    \item What does current evidence say about the duration of Long COVID symptoms, and are there predictors of recovery?
\end{enumerate}

\subsection*{Safety and Limitations}
\begin{enumerate}
    \setcounter{enumi}{16}
    \item What are the potential pitfalls in diagnosing Long COVID in a patient with multiple comorbidities?
    \item What are the limitations of current evidence on Long COVID treatment?
    \item How do you ensure patients with Long COVID receive equitable care?
    \item Patients are using nicotine patches to treat Long COVID, is that an effective treatment? Why would nicotine improve Long COVID symptoms?
\end{enumerate}

\section*{Appendix C: Retrieval architecture}
\textbf{Dense retrieval implementation.} For expert-curated corpuses (G-1, GS-4, R-110), we implemented a preprocessing pipeline for full-text document access and vector database construction. Full texts were retrieved via the PubMed Central (PMC) API when available; otherwise, PDFs were converted to plaintext using PyPDF with manual removal of non-content elements (headers, metadata, figures, numerical tables).
Documents were segmented using recursive character splitting with 1200-character chunks and 600-character overlap to preserve paragraph structure. Each chunk was embedded using OpenAI's text-embedding-3-small model (dimensionality: 1536) and indexed in a FAISS vector database.
During inference, queries were embedded using the same model, and the top-25 nearest neighbors (cosine similarity) were retrieved and concatenated with citation metadata for response generation via LangChain and GPT-4o.

\textbf{Hybrid sparse-dense retrieval for PubMed.} The PM corpus employed a two-stage retrieval process. Initially, GPT-4o generated PubMed search queries from input questions using single-shot prompting. PubMed's Best Match algorithm (BM25 + LambdaMART ranking) returned candidate documents \cite{fiorini2018best}. If fewer than 25 results were obtained, query generalization was iteratively applied until the threshold was met (Appendix A).
Full texts were retrieved via the PMC API (abstracts used when unavailable), then processed through the same chunking and dense retrieval pipeline described above, yielding the top-25 most relevant chunks for generation.

\begin{figure}[t]
    \centering
    \includegraphics[width=1\linewidth]{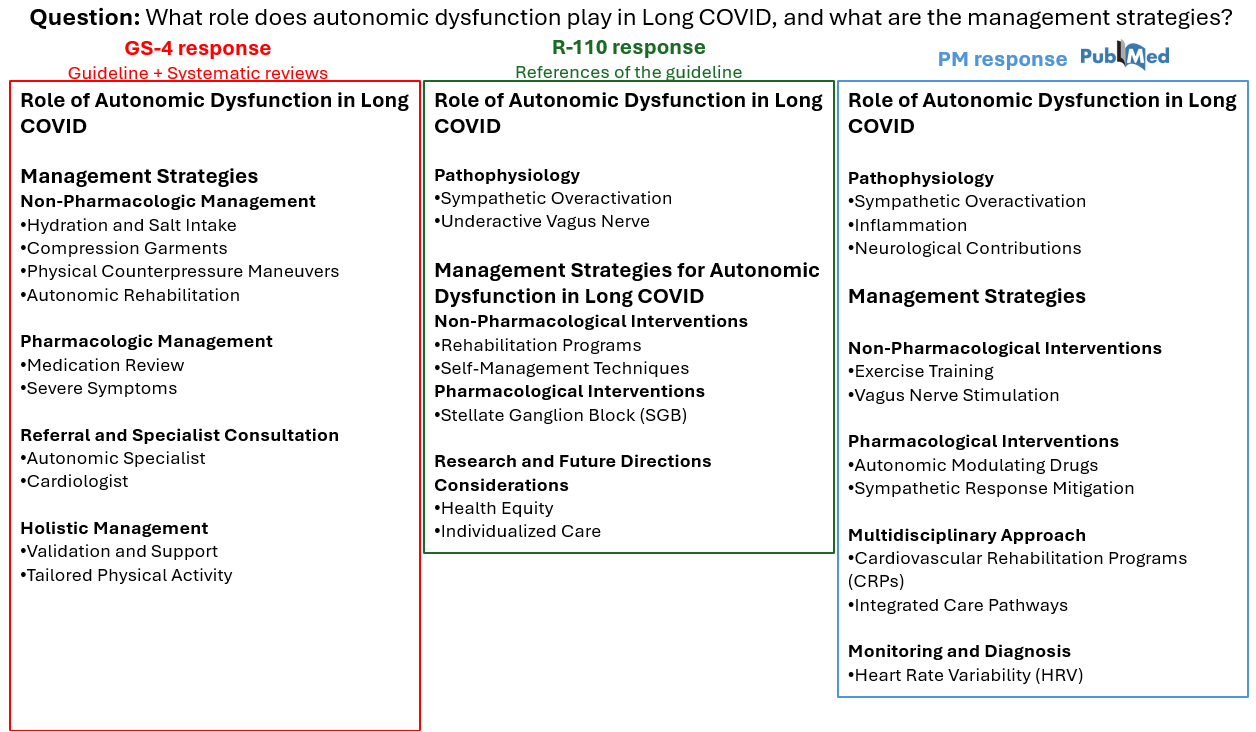}
    \caption{Comparison of answers to an example question from the LongCOVID-CQ dataset using the GS-4, R-110, and PM corpuses. Each response's section headers are included in the figure for conciseness.}
    \label{fig:Examples1}
\end{figure}

\section*{Appendix D: GS-4 response example}

We show an example response to the question ``What role does autonomic dysfunction play in Long COVID, and what are the management strategies?'' using the GS-4 corpus.

\subsubsection*{Role of Autonomic Dysfunction in Long COVID}
Autonomic dysfunction is a significant component of Long COVID, affecting a substantial number of patients. Studies have shown that a large proportion of individuals with Long COVID experience moderate to severe autonomic dysfunction, as indicated by high scores on the Composite Autonomic Symptom 31 (COMPASS-31) assessment (Fineberg\_39110819). Common manifestations include postural orthostatic tachycardia syndrome (POTS) and orthostatic intolerance, which can significantly impact daily functioning (Fineberg\_39110819).

\subsubsection*{Management Strategies}

\paragraph{Non-Pharmacologic Management}
\begin{itemize}
    \item \textbf{Hydration and Salt Intake:} Increase oral intake to 2.5–3.5 liters of fluid and 2.8–4 grams of sodium daily (AAPMRCompendium\_NP002).
    \item \textbf{Compression Garments:} Use waist-high stockings and/or abdominal binders to help manage orthostatic intolerance (AAPMRCompendium\_NP002).
    \item \textbf{Physical Counterpressure Maneuvers:} Techniques such as standing with legs crossed or sitting in a knee-chest position can mitigate symptoms (AAPMRCompendium\_NP002).
    \item \textbf{Autonomic Rehabilitation:} Engage in a personalized rehabilitation program, starting with supine core strengthening or recumbent activities like biking or swimming (AAPMRCompendium\_NP002).
\end{itemize}

\paragraph{Pharmacologic Management}
\begin{itemize}
    \item \textbf{Medication Review:} Consider weaning or de-prescribing medications that may exacerbate symptoms such as orthostatic intolerance, tachycardia, and hypotension (AAPMRCompendium\_NP002).
    \item \textbf{Severe Symptoms:} Medication management should be considered if symptoms are severe or persistent after non-pharmacologic measures. Intravenous fluids may be used sparingly during symptom flares (AAPMRCompendium\_NP002).
\end{itemize}

\paragraph{Referral and Specialist Consultation}
\begin{itemize}
    \item \textbf{Autonomic Specialist:} Referral to an autonomic specialist is recommended if non-pharmacologic and first-line pharmacologic management do not ameliorate symptoms, or if there is frequent syncope or a need for additional autonomic function assessment (AAPMRCompendium\_NP002).
    \item \textbf{Cardiologist:} If an autonomic specialist is unavailable, a general cardiologist may be consulted for management outside the scope of the treating clinician (AAPMRCompendium\_NP002).
\end{itemize}

\paragraph{Holistic Management}
Validation and Support: It is crucial to validate the patient's experience and provide reassurance about the reality of Long COVID symptoms, emphasizing that many patients can manage their symptoms and improve their quality of life with available treatment strategies (AAPMRCompendium\_NP002).

Tailored Physical Activity: Physical activity recommendations should be carefully tailored to avoid triggering post-exertional malaise (AAPMRCompendium\_NP002).

These strategies aim to address the complex and multifaceted nature of autonomic dysfunction in Long COVID, providing a comprehensive approach to management.

\section*{Appendix E: R-110 response example}

We show an example response to the question ``What role does autonomic dysfunction play in Long COVID, and what are the management strategies?'' using the R-110 corpus.

\subsubsection*{Role of Autonomic Dysfunction in Long COVID}
Autonomic dysfunction, particularly dysautonomia, plays a significant role in the pathophysiology of Long COVID. The autonomic nervous system (ANS) is crucial in managing acute hyperinflammation and endothelial dysfunction during viral infections like COVID-19. In Long COVID, a malfunctioning ANS is often marked by sympathetic hyperactivity and an underactive vagus nerve, which may lead to persistent inflammation and disrupt the balance between the sympathetic and parasympathetic systems (Pearson\_37711269). This imbalance is thought to contribute to the characteristic symptoms of Long COVID, such as fatigue, cognitive issues, and cardiovascular dysautonomia (Rubin\_32965460).

\subsubsection*{Management Strategies for Autonomic Dysfunction in Long COVID}

\paragraph{Non-Pharmacological Interventions}
\begin{itemize}
    \item \textbf{Rehabilitation Programs:} Cardiovascular rehabilitation and telerehabilitation exercise programs have been shown to improve cognitive deficits in Long COVID patients (Sathyamoorthy\_38206585).
    \item \textbf{Self-Management Techniques:} Patients often resort to self-prescribed modalities, including dietary modifications and over-the-counter supplements. However, these should be pursued with caution and under professional guidance to avoid risks such as drug interactions (Sweis\_38541324).
\end{itemize}

\paragraph{Pharmacological Interventions}
\textbf{Stellate Ganglion Block (SGB):} This intervention targets the sympathetic nervous system and has shown promise in alleviating symptoms of dysautonomia in Long COVID by increasing cerebral and regional blood flow, potentially resetting ANS function (Khan\_36628048).

\paragraph{Research and Future Directions}
The PASC Collaborative highlights the need for further studies to determine biomarkers and appropriate diagnostic and therapeutic approaches for post-COVID autonomic dysfunction (Blitshteyn\_36169154).

Clinical trials are ongoing to evaluate the effectiveness of various pharmacological interventions, including immunotherapy and other novel treatments (Reis\_NP020).

\paragraph{Considerations}
\begin{itemize}
    \item \textbf{Health Equity:} Special considerations are necessary for populations with pre-existing autonomic disorders and those facing health disparities, as these groups may experience a higher burden of Long COVID symptoms (Blitshteyn\_36169154).
    \item \textbf{Individualized Care:} Given the variability in symptoms and responses to treatment, personalized management plans are essential. This includes tailored investigations and specialized consultations based on individual symptoms (Sweis\_38541324).
\end{itemize}

In summary, autonomic dysfunction is a key component of Long COVID, and while management strategies are still evolving, a combination of rehabilitation, self-management, and targeted pharmacological interventions shows potential. Ongoing research is crucial to develop evidence-based guidelines and improve patient outcomes.

\section*{Appendix F: PM response example}

We show an example response to the question ``What role does autonomic dysfunction play in Long COVID, and what are the management strategies?'' using the PM corpus.

\subsubsection*{Role of Autonomic Dysfunction in Long COVID}
Autonomic dysfunction is a significant component of Long COVID, manifesting as dysautonomia, which includes symptoms such as fatigue, arrhythmias, and hypertension. This dysfunction is linked to the cardiovascular and neurological aspects of Long COVID (Allendes\_36743679). Dysautonomia in Long COVID can present as postural orthostatic tachycardia syndrome (POTS), characterized by an exaggerated sympathetic response and orthostatic intolerance (Astin\_36412084; Ewing\_40254579).

\subsubsection*{Pathophysiology}
\begin{itemize}
    \item \textbf{Sympathetic Overactivation:} Chronic activation of the sympathetic nervous system can lead to arrhythmogenesis, orthostatic hypotension, and cognitive decline (Allendes\_36743679).
    \item \textbf{Inflammation:} Microinflammation in autonomic nerve fibers, termed ``post-COVID Guillan-Barré syndrome'' (PCGBS), is a specific type of dysautonomia observed in Long COVID (Allendes\_36743679).
    \item \textbf{Neurological Contributions:} Dysautonomia and brain fog suggest neurological involvement, with inflammation playing a critical role in the pathogenesis (Astin\_36412084).
\end{itemize}

\subsubsection*{Management Strategies}

\paragraph{Non-Pharmacological Interventions}
\begin{itemize}
    \item \textbf{Exercise Training:} Exercise has shown potential in improving cardiovascular and autonomic function. It may enhance vascular endothelial function and autonomic balance, making it a promising complementary strategy for Long COVID patients with dysautonomia (Allendes\_36743679).
    \item \textbf{Vagus Nerve Stimulation:} Techniques that stimulate the vagus nerve, such as breath retraining, HRV biofeedback, and yoga, can help reduce sympathetic activity and increase vagal tone (Astin\_36412084; Ewing\_40254579).
\end{itemize}

\paragraph{Pharmacological Interventions}
\begin{itemize}
    \item \textbf{Autonomic Modulating Drugs:} Medications like ivabradine, beta-blockers, and midodrine can be useful in managing symptoms of dysautonomia in Long COVID patients (Ewing\_40254579).
    \item \textbf{Sympathetic Response Mitigation:} beta-receptor blockers may help mitigate symptoms from exaggerated sympathetic responses (Astin\_36412084).
\end{itemize}

\paragraph{Multidisciplinary Approach}
\begin{itemize}
    \item \textbf{Cardiovascular Rehabilitation Programs (CRPs):} These programs are interdisciplinary interventions that can stimulate the parasympathetic nervous system, improve cardiorespiratory fitness, and reduce cardiovascular risk factors (Allendes\_36743679).
    \item \textbf{Integrated Care Pathways:} Combining rehabilitation, drug interventions, and research into healthcare access can provide comprehensive management for Long COVID (Astin\_36412084).
\end{itemize}

\paragraph{Monitoring and Diagnosis}
\textbf{Heart Rate Variability (HRV):} HRV analysis is a suitable tool for diagnosing post-COVID dysautonomia and guiding treatment strategies (Allendes\_36743679).

In summary, autonomic dysfunction plays a crucial role in Long COVID, with management strategies focusing on both non-pharmacological and pharmacological interventions to restore autonomic balance and improve patient outcomes.

\section*{Appendix G: LLM-as-a-judge comparison}
The following was the output of the LLM-as-a-judge based on comprehensiveness between GS-4 (1) vs. R-110 (2).

\begin{quote}
Answer 1 provides a more comprehensive overview of both the role of autonomic dysfunction in Long COVID and the management strategies. It details specific non-pharmacologic and pharmacologic interventions, including hydration, compression garments, and medication review, as well as the importance of specialist consultation and holistic management. Answer 2, while informative, focuses more on the theoretical aspects and future research directions, lacking the detailed management strategies provided in Answer 1.
\end{quote}

\section*{Appendix H: R-110 Corpus Preprocessing}

Two references of the AAPM\&R compendium were excluded from the R-110 corpus due to infeasibility of preprocessing. Reference 48's full text was not available in English, only in German. Reference 71 was a website compiling links to social security regulations, which would require comprehensive web-scraping to parse with limited relevance to our clinical questions. Both citations are shown below. 

48 Laskowski NM, Brandt G, Paslakis G. Geschlechtsspezifische Unterschiede und Ungleichheiten der COVID-19 Pandemie: Eine Synthese systematischer Reviews unter Einbeziehung sexueller und geschlechtlicher Minderheiten. Psychother Psychosom Med Psychol. 2024;74(2):57-69. doi:10.1055/a-2228-6244

71 Social Security Administration. Code of Federal Regulations: PART 404 FEDERAL OLD-AGE, SURVIVORS AND DISABILITY INSURANCE (1950); U.S. Social Security Administration, Revised as of April 1, 2023; accessed via \url{https://www.ssa.gov/OP\_Home/cfr20/404/404-0000.htm} on August 4, 2024.

At the time our research was conducted, the latest version of the AAPM\&R compendium had 113 references total. Future published versions may have a different number of references due to subsequent author revisions.
\end{document}